\documentclass[11pt]{article}

\usepackage[top=50mm, bottom=50mm, left=50mm, right=50mm]{geometry}
\usepackage{lineno}
\usepackage{amssymb}
\usepackage{amsmath}
\usepackage{amsthm}
\usepackage{epsfig}
\usepackage{graphicx}
\usepackage{graphics}
\usepackage{float}
\usepackage{subfigure}
\usepackage{multirow}
\usepackage{color}
\usepackage{lineno}
\usepackage{fullpage}
\usepackage[normalem]{ulem} 
\usepackage{makeidx}
\usepackage{xspace}
\usepackage{wrapfig}
\usepackage{hyperref}
\usepackage{algorithm}
\usepackage{algpseudocode}
\usepackage[nameinlink]{cleveref}
\usepackage{makecell}
\usepackage{booktabs}

\begin{document}
\title{Physics-Informed Koopman Network}

\author{
{Yuying Liu$^{\dag}$ Aleksei Sholokhov$^{\dag}$ Hassan Mansour$^{\S}$  Saleh Nabi$^{\S}$}\\[.1in]
{$^{\dag}$ {\small Department of Applied Mathematics, University of Washington, Seattle, WA 98105}}\\
{$^{\S}$ {\small Mitsubishi Electric Research Laboratories, Cambridge, MA 02139}}\\
}
\maketitle

\begin{abstract}
    Koopman operator theory is receiving increased attention due to its promise to linearize nonlinear dynamics. Neural networks that are developed to represent Koopman operators have shown great success thanks to their ability to approximate arbitrarily complex functions. However, despite their great potential, they typically require large training data-sets either from measurements of a real system or from high-fidelity simulations. In this work, we propose a novel architecture inspired by physics-informed neural networks, which leverage automatic differentiation to impose the underlying physical laws via soft penalty constraints during model training. We demonstrate that it not only reduces the need of large training data-sets, but also maintains high effectiveness in approximating Koopman eigenfunctions.
\end{abstract}

\section{Introduction}
Nonlinear dynamical systems give rise to a rich diversity of complex phenomena. Dealing with nonlinearity is the central task of many areas of science and engineering such as climate science \cite{lorenz1956empirical}, neuroscience \cite{brunton2016extracting}, ecology \cite{clark2020nonlinear}, finance \cite{mann2016dynamic}, and epidemiology \cite{proctor2015discovering}. A classic yet popular view of dynamical systems is geometric, where the behavior of multiple trajectories can be simultaneously studied in a qualitative manner. However, the geometry in state space becomes more complex when the dynamics become nonlinear and high dimensional, making the system hard to analyze, predict and control~\cite{mezic2020koopman}.

In 1931, Koopman introduced the operator-theoretic perspective of dynamical systems \cite{koopman1931hamiltonian}, complementing the traditional geometric perspectives. In this framework, a Koopman operator is defined which acts on observation functions (observables) in an appropriate function space. Under the action of this operator, the evolution of the observables are linear although the function space may be infinite-dimensional. As a consequence, approximating the Koopman operator and seeking its eigenfunctions become a key to linearize the nonlinear dynamics \cite{mezic1994geometrical,mezic2005spectral,brunton2021modern}.

The leading computational method for approximating the Koopman operator is dynamic mode decomposition (DMD) \cite{rowley2009spectral}.
Using the snapshots of state measurements of a system, the DMD algorithm seeks the best linear operator that approximately advances these states. There are many variants of DMD, however, most of them require an a priori, judicious selection of the observables, and there is no guarantee that these observables span an invariant Koopman subspace \cite{kutz2016dynamic}. Another limitation is that a typical application of DMD starts from a single trajectory and thererfore the approximation of the Koopman operator is only restricted to these measurements, leading to failures of correctly predicting unseen trajectories for non-ergodic systems \cite{arbabi2017ergodic}. 

To address these challenges, neural networks were proposed to approximate the Koopman operator. In terms of architecture, autoencoder is the most commonly used building block. 
Furthermore, a linearity constraint is added to the loss function to approximate the Koopman operator \cite{takeishi2017learning,morton2018deep,lusch2018deep,gin2021deep}.
The neural network approach generalizes well mainly due to three reasons: (i) it automatically selects the observables \cite{takeishi2017learning}, (ii) it can approximate arbitrarily complex functions \cite{hornik1989multilayer}, and (iii) trained with multiple trajectories, leading to a good interpolation of the dynamics. A successfully trained neural network can identify coordinate transformations that make strongly nonlinear dynamics approximately linear and, therefore, enabling linear prediction, estimation, and control. However, one would need to acquire a large enough data-set to successfully train a neural network which is neither efficient nor practical.

On another line of research, physics-informed neural networks (PINNs)\cite{raissi2019physics} were introduced in 2019. They can seamlessly integrate the measurement data and physical governing laws by penalizing the residuals of the differential equation in the loss function using automatic differentiation. This approach alleviates the need for a large amount of data by assimilating the knowledge of the equations into the training process. However,
(i) PINNs can only solve one solution instance at a time, and (ii) the solution is not accurate outside the training time horizon.

In this work, we propose physics-informed Koopman networks (PIKNs) which combines the strengths of both PINNs and autoencoder-based Koopman networks. More specifically, by incorporating the knowledge of dynamics, we reduce the need for large training data-sets for identifying Koopman eigenvalues and eigenfunctions. Moreover, since the network performs (Koopman) operator learning, it enables the model to predict beyond the training horizon, and also to be used for compressed sensing, estimation, and control. 

The paper is organized as follows. In Section~\ref{sec:method}, we briefly introduce Koopman operator theory and our methodology. Then we mention some related works in Section~\ref{sec:related_work} and highlight the connections and differences. Our approach is then tested on several benchmark problems in Section~\ref{sec:exp}. In Section~\ref{sec:conclusion}, we conclude and discuss future directions. 

\section{Method}
\label{sec:method}
\subsection{Koopman operator theory}
We begin with introducing an autonomous ordinary differential equation
\begin{equation}
    \label{eq:ode}
    \frac{d}{dt}\mathbf{x}(t)=\mathbf{f}(\mathbf{x}(t))
\end{equation}
with $\mathbf{x}\in \mathcal{X} \subseteq \mathbb{R}^n$. We define the time-$t$ flow map operator $\mathbf{F}^t: \mathcal{X} \rightarrow \mathcal{X}$ as 
\begin{equation}
    \mathbf{x}(t_0+t) = \mathbf{F}^t(\mathbf{x}(t_0))
\end{equation}
In 1931, B.O.Koopman \cite{koopman1931hamiltonian} provided an alternative description for dynamical systems in terms of evolution of functions of possible measurements $\mathbf{y} = g(\mathbf{x})$. The function $g: \mathcal{X} \rightarrow \mathbb{C}$ is called a measurement function and it belongs to some set of functions $\mathcal{G(\mathcal{X})}$. This set is often not defined \textit{a priori}, and Hilbert spaces such as $L^2(\mathcal{X}, d\mu)$ or reproducing kernel Hilbert spaces (RKHS) are common choices \cite{das2020koopman,mezic2020koopman}. In all cases, however, $\mathcal{G(\mathcal{X})}$ is of significantly higher dimension than $\mathcal{X}$, so we are trading dimensionality for linearity. \\
The family of Koopman operators $\mathcal{K}^t: \mathcal{G(\mathcal{X})} \rightarrow \mathcal{G(\mathcal{X})}$, parameterized by $t$ are given by
\begin{equation}
    \label{eq:koopman}
    \mathcal{K}^t g(\mathbf{x}) = g(\mathbf{F}^t(\mathbf{x}))
\end{equation}
One can check that $\mathcal{K}^t$ is linear. In general it is infinite-dimensional, and constructing finite-dimensional representations of Koopman operator remains an open question.

If the dynamics in Equation~\ref{eq:ode} is sufficiently smooth, one can also define the infinitesimal generator $\mathcal{L}$ of the Koopman operator family as 
\begin{equation}
    \label{eq:generator}
    \mathcal{L}g := \lim_{t\rightarrow 0} \frac{\mathcal{K}^t g - g}{t} = \lim_{t\rightarrow 0}\frac{g\circ \mathbf{F}^t - g}{t}
\end{equation}
From the definition, we can easily see
\begin{equation}
    \mathcal{L}g(\mathbf{x}(t)) = \lim_{\tau \rightarrow 0} \frac{g(\mathbf{x}(t+\tau))-g(\mathbf{x}(t))}{\tau} = \frac{d}{dt}g(\mathbf{x}(t))
\end{equation}
The generator $\mathcal{L}$ is sometimes referred to as the Lie operator: it is the Lie derivative of $g$ along the vector field $\mathbf{f}(\mathbf{x})$ when the dynamics is given by Equation~\ref{eq:ode}. On the other hand, we also have
\begin{equation}
    \frac{d}{dt}g(\mathbf{x}(t)) = \nabla g \cdot \frac{d}{dt}\mathbf{x}(t) = \nabla g \cdot \mathbf{f}(\mathbf{x}(t))
\end{equation}
Therefore, we conclude
\begin{equation}
    \label{eq:lie_derivative}
    \mathcal{L}g = \nabla g \cdot \mathbf{f}
\end{equation}
Equation \ref{eq:lie_derivative} will be the key for the implementation of PIKN. 

Applied Koopman analysis seeks key measurement functions that behave linearly in time, and the eigenfunctions of the Koopman operator are functions that exhibit such behaviour \cite{brunton2021modern,mezic2022numerical}. A Koopman eigenfunction $\varphi(\mathbf{x})$ corresponding to an eigenvalue $\lambda^t$ satisfies
\begin{equation}
    \label{eq:eigenfunc}
    \mathcal{K}^t \varphi(\mathbf{x}) = \lambda^t \varphi(\mathbf{x})
\end{equation}
 It is straightforward to show that Koopman eigenfunctions $\varphi(\mathbf{x})$ that satisfies Equation~\ref{eq:eigenfunc} for $\lambda^t \neq 0$ are also eigenfunctions of the Lie operator, although with a different eigenvalue $\mu = \log(\lambda^t)/t$.

 Once we have a set of eigenfunctions $\{\varphi_1, \varphi_2, \cdots, \varphi_M\}$, observables that can be formed as a linear combination of these eigenfunctions, i.e., $g \in span\{\varphi_k\}_{k=1}^{M}$ have a particularly simple evolution under the Koopman operator
\begin{equation}
    \label{eq:linear_reconstruct}
    g(\mathbf{x}) = \sum_{k=1}^M c_k\varphi_k(\mathbf{x})\ \ \Longrightarrow \ \ \mathcal{K}^t g(\mathbf{x}) = \sum_{k=1}^M c_k \lambda_k^t \varphi_k(\mathbf{x})
\end{equation}
This also implies $span\{\varphi_k\}_{k=1}^{M}$ is an invariant subspace under the Koopman operator $\mathcal{K}^t$ and can be viewed as the new coordinates on which the dynamics evolve linearly. 

Although the promise of Koopman operator theory is tempting, there are certain challenges. For example,
\begin{itemize}
    \item The Koopman operator of a system may not admit a discrete spectrum. Certain systems fundamentally fail to fit into this framework \cite{lusch2018deep,koopman1932dynamical}. 
    \item Asymptotic methods can be used to approximate certain eigenfunctions for simple dynamics (e.g., polynomial nonlinear dynamics), however, there is no analytical procedure to seek for the eigen-pairs of Koopman operator in general.
    \item Some computational methods (e.g., DMD) can be used to approximate eigenfunctions of Koopman operator, but the approximation is only restricted to those measurements leading to spurious modes. Moreover, the identified basis may not span a Koopman invariant subspace.
\end{itemize}

Since our ultimate goal is to study nonlinear dynamical systems using linear theory, we do not need to restrict ourselves to Equation~\ref{eq:linear_reconstruct}. Following \cite{lusch2018deep, gin2021deep}, we can generalize it as
\begin{equation}
    \label{eq:nonlinear_reconstruct}
    \begin{split}
        g(\mathbf{x}) &= \psi(\varphi_1(\mathbf{x}), \varphi_2(\mathbf{x}), \dots, \varphi_M(\mathbf{x}); \mathbf{\omega}) \\
        &\hspace{2cm} \big\Downarrow \\
        \mathcal{K}^t g(\mathbf{x}) &= \psi(\lambda_1^t\varphi_1(\mathbf{x}), \lambda_2^t\varphi_2(\mathbf{x}), \dots, \lambda_M^t\varphi_M(\mathbf{x}); \mathbf{\omega})
    \end{split}
\end{equation}
where $\psi$ is an arbitrary transformation parameterized by $\mathbf{\omega}$.

\subsection{Auto-encoder based architecture}
An auto-encoder is a special type of neural network which is particularly suitable for our application and widely used in the literature \cite{lusch2018deep, gin2021deep}.
The encoder $\phi$ learns the representation of the relevant Koopman eigenfunctions, which provides intrinsic coordinates that linearize the dynamics, and the decoder $\psi$ seeks an inverse transformation to reconstruct the original measurements. One may hope that if we define $\phi: \mathbf{x} \rightarrow (\varphi_1(\mathbf{x}), \varphi_2(\mathbf{x}), \dots, \varphi_M(\mathbf{x}))^T $, then up to a constant, the encoder learns this transformation $\phi$ and the decoder learns the $\psi$ as shown in Equation~\ref{eq:nonlinear_reconstruct}. (Alternatively, if we specify a linear decoder, then the learning regime would correspond to Equation~\ref{eq:linear_reconstruct}.) 

Within the auto-encoder's latent space, the dynamics is constrained to be linear. Therefore in previous works, a squared matrix $K$ is often used to drive the evolution of the dynamics.
Most often there is no invariant, finite-dimensional Koopman subspace that captures the evolution of all the measurements, then that matrix $K$ will only be an approximation of the true underlying linear operator. 

There are different ways to train the auto-encoder architecture in the literature \cite{takeishi2017learning, morton2018deep, lusch2018deep, gin2021deep, azencot2020forecasting}, however, they all require a large amount of measurement data. Normally, the training data-set $X$ is arranged as a 3D tensor, with its dimensions to be (i) number of sequences (with different initial states), (ii) number of snapshots, and (iii) dimensionality of the measurements, respectively. Then the constraint of linear dynamics can be enforced by a loss term resembling $\|\phi(\mathbf{x}_{n+1}) - K\phi(\mathbf{x}_n)\|$, or more generally, linearity is enforced over multiple steps $\|\phi(\mathbf{x}_{n+p}) - K^p\phi(\mathbf{x}_n)\|$, generating recurrencies in the neural network architecture. We will see, however, such large data-sets are not necessary (but beneficial) for PIKN.

\subsection{Physics-informed Koopman network}
In physical sciences, data is scarce while governing equations are available in literature. In physics-informed Koopman networks (PIKNs), we aim to leverage such knowledge of the dynamics, e.g.,  of Equation~\ref{eq:lie_derivative}, to enforce the linearity constraint. The basic idea is to train the network by  minimizing the quantity $\|\nabla \varphi_k(\mathbf{x}) \cdot \mathbf{f} - \mu_k \varphi_k(\mathbf{x})\|, \ \ \forall k = 1, 2, 3, \cdots, M$. More generally, a squared matrix $L$ is used to approximate the Lie operator $\mathcal{L}$, which in turn is related to the Koopman operator, and we minimize $\|L\phi(\mathbf{x}) - \nabla \phi(\mathbf{x}) \cdot \mathbf{f}\|$. Finding the eigenvalue and eigenfunction pairs of the Lie operator corresponds to performing eigendecomposition to the matrix $L$.

\subsubsection{For ODE}
We first consider an ordinary differential equation of the form in Equation~\ref{eq:ode}. We start from sampling a set of collocation points $X := \{\mathbf{x}_1, \mathbf{x}_2, \cdots, \mathbf{x}_N\}$. This set of collocation points does not need to come from any trajectories of the dynamics but they can be sampled randomly, avoiding a bulk of simulations or measurement data collections. 

The objective of the network is to identify a few key coordinates $\mathbf{z}=\phi(\mathbf{x})$ spannned by a set of Koopman eigenfunctions $\varphi_k(\mathbf{x}): \mathcal{R}^n \rightarrow \mathcal{R}, \ k=1,2,\cdots, M$ along with a dynamical system $\dot{\mathbf{z}}=L\mathbf{z}$. Objective function is 
\begin{equation}
    \mathcal{J}_{linear} = \frac{1}{N}\sum_{i=1}^N (\omega_1\|L \phi(\mathbf{x}_i) - \nabla \phi(\mathbf{x}_i) \cdot \mathbf{f}(\mathbf{x}_i)\|^2 + \omega_2\|\mathbf{x}_i - C\mathbf{z}_i\|^2)
\end{equation}
if the decoder is linear (where $C$ represents the reconstruction coefficients), or for generic decoder
\begin{equation}
    \mathcal{J}_{nonlinear} = \frac{1}{N}\sum_{i=1}^N (\omega_1\|L \phi(\mathbf{x}_i) - \nabla \phi(\mathbf{x}_i) \cdot \mathbf{f}(\mathbf{x}_i)\|^2 + \omega_2\|\mathbf{x}_i - \psi(\mathbf{z_i})\|^2)
\end{equation}
Here, $\omega_1$ and $\omega_2$ are the weights for each loss term. The first term encourages linear dynamics within the latent space and the second term makes it a valid auto-encoder. One can further diagonalize $L$ such that the diagonal elements approximate the Koopman eigenvalues and the corresponding outputs of the encoder approximate the Koopman eigenfunctions, respectively. But this constraint is not necessary as it is equivalent to performing eigendecomposition of a general-structured $L$ after training. Once trained, it can be used for state predictions beyond the time horizon used in training.

\subsubsection{For PDE}
For application to partial differential equations of the form
\begin{equation}
    \label{eq:pde}
    \mathbf{u}_t = \mathbf{f}(\mathbf{u}, \mathbf{u}_\mathbf{x}, \dots)
\end{equation}
similar to ODEs, the goal is to seek coordinates $\mathbf{v}=\phi(\mathbf{u})$ that has linear evolution $\dot{\mathbf{v}}=L\mathbf{v}$ and can be used to reconstruct the original measurements $\hat{\mathbf{u}}=\psi(\mathbf{v})$. The main difference is, however, the input and output of the network are functions of spatio-temporal variables $\mathbf{u}(\mathbf{x, t})$ instead of the temporal variables $\mathbf{x}(t)$ as in the ODE cases. Therefore, we need to sample a set of "collocation points" in an appropriate function space where $\mathbf{u}_t$ can be cheaply evaluated. We provide examples of such suitable function families in the following chapters. 

\subsection{Data integration}

Like PINNs, we can seamlessly integrate information from measurement data. Suppose we have snapshots of measurements $X_{data} := \{\mathbf{x}(t_0), \mathbf{x}(t_1), \cdots, \mathbf{x}(t_p)\}$ for an ODE system or $U_{data} := \{\mathbf{u}(\mathbf{x}, t_0), \mathbf{u}(\mathbf{x}, t_1), \cdots, \mathbf{u}(\mathbf{x}, t_p)\}$ for a PDE, by adding extra loss terms
\begin{equation}
    \label{eq:Jdata}
    \begin{split}
        \mathcal{J}_{data} &= \frac{1}{p}\sum_{j=0}^p (\omega_3\|e^{L\Delta t_j}\phi(\mathbf{x}(t_0)) - \phi(\mathbf{x}(t_j))\|^2 + \omega_4\|\mathbf{x}(t_j) - \psi(\mathbf{z}(t_j))\|^2) \ \ \text{(for ODE)}\\
        \mathcal{J}_{data} &= \frac{1}{p}\sum_{j=0}^p (\omega_3\|e^{L\Delta t_j}\phi(\mathbf{u}(\mathbf{x}, t_0)) - \phi(\mathbf{u}(\mathbf{x}, t_j))\|^2 + \omega_4\|\mathbf{u}(\mathbf{x}, t_j) - \psi(\mathbf{v}(\mathbf{u}(\mathbf{x}, t_j)))\|^2) \ \ \text{(for PDE)}\\
    \end{split}
\end{equation}
we can penalize the network predictions that do not match the real measurements, where $\Delta t_j = t_j - t_0, \ \ \forall j = 1, 2, \cdots, p$.
Again, the first term is the linearity loss and the second term is the reconstruction loss. It should be noted that Eq. \ref{eq:Jdata} is consistent with previous literature \cite{takeishi2017learning,morton2018deep,lusch2018deep} on using autoencoders to find approximation of Koopman eigenfunctions and can be seen as a baseline on how physics-informed loss improve the performance of PIKN.


\section{Related Work}
\label{sec:related_work}
\subsection{Dynamic Mode Decomposition}
Dynamic mode decomposition (DMD), which was originally introduced by Schmid\cite{schmid2010dynamic}, is the leading computational method to approximate the Koopman operator from data\cite{kutz2016dynamic, tu2013dynamic}. Rowley et al. were the first to establish the connection between DMD and Koopman operator theory\cite{rowley2009spectral}. One of the major advantages of DMD is its simple formulation in terms of linear regression. For this reason, many methodological innovations have been introduced, for example, Jovanovic et al.\cite{jovanovic2014sparsity} uses sparsity promoting optimzation to identify the fewest DMD modes; \cite{bistrian2017randomized, erichson2019randomized} accelerate DMD using randomized linear algebra; extended DMD\cite{williams2015data} suggests to include nonlinear measurements; higher order DMD\cite{le2017higher} acts on delayed coordinates and generates more complex models; multiresolution DMD\cite{kutz2016multiresolution} deals with multiscale systems that exhibit transient or intermittent dynamics; Proctor et al.\cite{proctor2016dynamic} extended the algorithm to disambiguate the natural dynamics and actuation; algorithms include total least-squares DMD\cite{hemati2017biasing}, forward-backward DMD\cite{dawson2016characterizing} and variable projection\cite{askham2018variable} improve the performance of DMD over noise sensitivity. These methods are now widely applied to many fields in science and engineering such as fluid dynamics and heat trasnfer\cite{bagheri2013koopman, basley2013space, bellani2011experimental, mizuno2011investigation,kalur2021robust,kramer2017sparse}, epidemiology\cite{proctor2015discovering}, neuroscience\cite{brunton2016extracting}, finance\cite{mann2016dynamic}, plasma physics\cite{taylor2018dynamic}, robotics\cite{berger2014dynamic, berger2015estimation} and video processing\cite{grosek2014dynamic, erichson2019compressed, bi2018dynamic}. For a more comprehensive review, one can refer to \cite{Brunton2019book}.

\subsection{Deep Learning for Linear Embedding}
Hand-crafted basis functions or measurements from DMD sometimes fail to represent the complex Koopman eigenfunctions. Neural networks turn out to be more effective in approximating them, leading to linear embedding of the nonlinear dynamics\cite{takeishi2017learning, li2017extended, lusch2018deep, yeung2019learning, azencot2020forecasting}. They have achieved great successes in long-term dynamic predictions\cite{lange2020fourier}, fluid control\cite{morton2018deep} and also recently be extended to account for uncertainties\cite{morton2019deep}, modeling PDEs\cite{gin2021deep} and jointly used with optimal control\cite{han2020deep, al2021deep}. There are also innovations on the side of neural network architectures, for example, neural ODEs are used for dictionary learning\cite{terao2021extended} and graphical neural networks are used to learn compositional Koopman operators\cite{li2019learning}. However, all the listed works are purely data-driven and do not address the issue of data efficiency.

\subsection{Physics-Informed Neural Network}
Physics-informed neural networks (PINNs) were first proposed in\cite{raissi2017physics, raissi2019physics} and have received lots of attention due to its flexibility to integrate measurement data and the physics (governing equations). They have been applied to various classes of PDEs\cite{pang2019fpinns, fang2019physics, zhang2020learning} and extended to deal with uncertainties\cite{yang2019adversarial, zhang2019quantifying, zhu2019physics, sun2020physics, yang2021b}. Another line of research of PINNs focus on its training and performance. For example, domain decomposition is considered in some variations of PINNs\cite{kharazmi2021hp, jagtap2020conservative, jagtap2020extended}, leading to parallel implementations\cite{shukla2021parallel, hennigh2021nvidia}; multi-fidelity framework\cite{meng2020composite}, dynamic weights of the loss function\cite{wang2021understanding} and hard constraints\cite{lu2021physics} have also been thoroughly studied, in order to achieve stable training results. Theoretical insights into the convergence of PINNs are presented in\cite{shin2020convergence, mishra2020estimates, wang2022and}. Recently, PINNs have also been jointly used with DeepONets\cite{wang2021learning}, entering the realm of operator learning. Similar to our work, the key motivation of using PINNs there is to eliminate the need of large data-sets for training DeepONets, and to achieve this we use automatic differentiation in evaluation of the loss function related to \ref{eq:lie_derivative}.

\section{Experiments}
\label{sec:exp}
\subsection{Simple nonlinear system with discrete spectrum}
    \begin{figure}[t!]
        \centering
        \includegraphics[width=\textwidth]{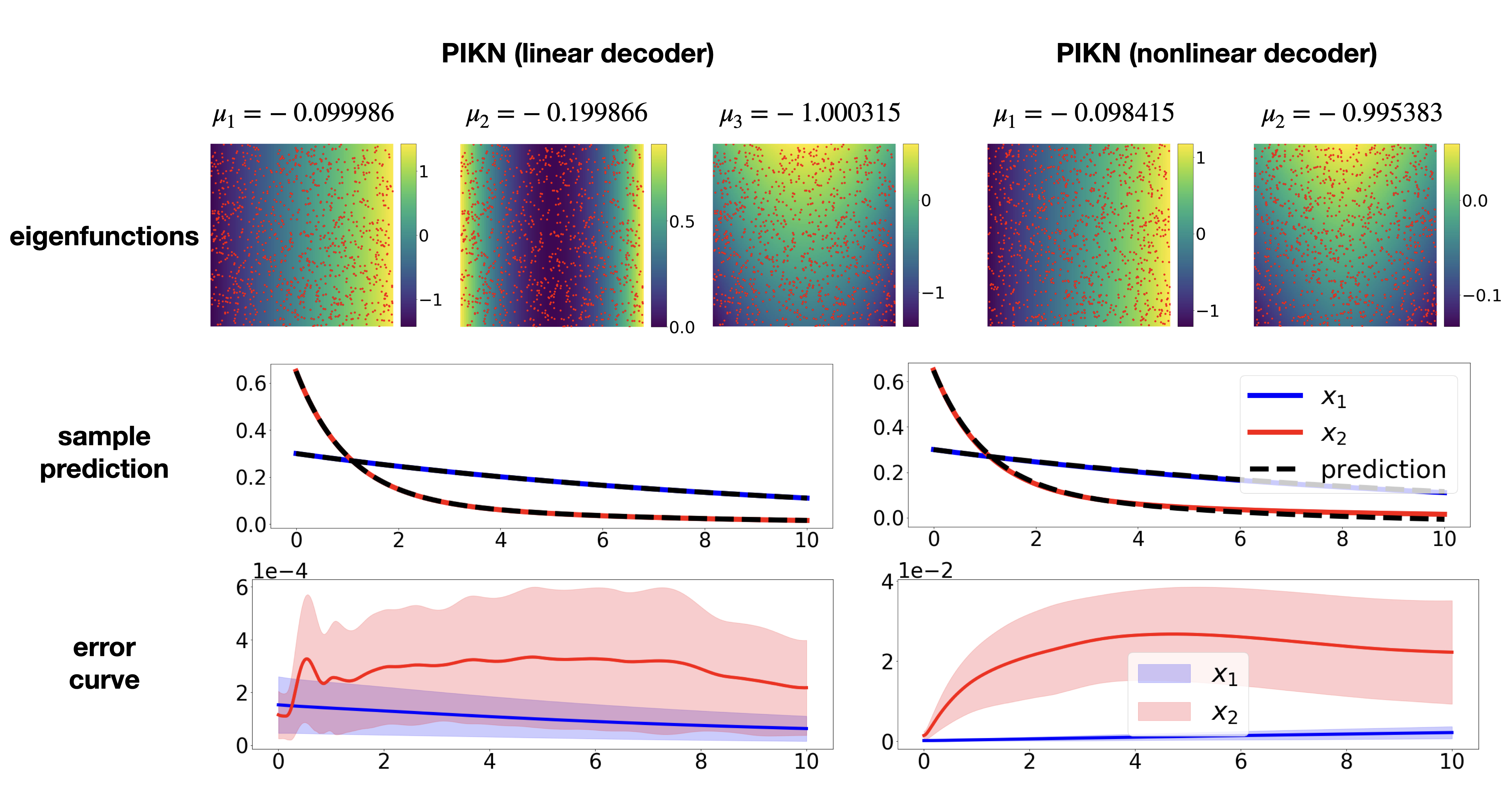}
        \vspace{-.2in}
        \caption{PIKN results using autoencoder with linear (left column) and nonlinear (right column) decoder network. The first row shows the eigenvalues and eigenfunctions identified by proposed PIKNs where the red dots represent the collocation points for training; the second row shows a $10000$-step forward prediction of the dynamics starting from a initial state $(x_1, x_2)=(0.3, 0.65)$ and with $\Delta t=0.001$; the third row visualizes the mean absolute error over $1000$ different trajectories with initial conditions uniformly sampled from $[-1, 1]\times[-1, 1]$, the shaded region covers one standard deviation away from the mean error.}
        \label{fig:nonlinear_ode}
    \end{figure}
    
    First, we consider a simple nonlinear system with a single fixed point and a discrete eigenvalue spectrum:
    \begin{equation}
        \label{eq:simpleODE}
        \begin{split}
            \dot{x_1} &= \mu x_1 \\
            \dot{x_2} &= \lambda (x_2 - x_1^2)
        \end{split}
    \end{equation}
    For $\lambda < \mu < 0$, the system exhibits a slow attracting manifold given by $x_2 = x_1^2$. In our experiment, we use $\mu=-0.1$ and $\lambda=-1$. This example is simple and widely adopted as a benchmark for Koopman/DMD related algorithms because it's possible to explicitly define a three-dimensional Koopman invariant subspace (spanned by Koopman eigenvalue and eigenfunction pairs) that contains the state variables $x_1$ and $x_2$:
    \begin{equation}
        \begin{split}
            \frac{d}{dt} 
                \begin{bmatrix}
                    \varphi_{\mu} \\
                    \varphi_{2\mu} \\
                    \varphi_{\lambda}
                \end{bmatrix} &=
                \begin{bmatrix}
                    \mu & 0 & 0 \\
                    0 & 2\mu & 0 \\
                    0 & 0 & \lambda
                \end{bmatrix}
                \begin{bmatrix}
                    \varphi_{\mu} \\
                    \varphi_{2\mu} \\
                    \varphi_{\lambda}
                \end{bmatrix} \\
            \begin{bmatrix}
                x_1 \\
                x_2 
            \end{bmatrix} &= 
                \begin{bmatrix}
                    1 & 0 & 0 \\
                    0 & b & 1 \\
                \end{bmatrix}
                \begin{bmatrix}
                    \varphi_{\mu} \\
                    \varphi_{2\mu} \\
                    \varphi_{\lambda}
                \end{bmatrix}
        \end{split}
    \end{equation}
    where $\phi(\mathbf{x}) = [\varphi_{\mu}, \varphi_{2\mu}, \varphi_{\lambda}] = [x_1, x_1^2, x_2-bx_1^2]$ and $b = \frac{\lambda}{\lambda-2\mu}$. If nonlinear decoding is allowed, a two-dimensional Koopman invariant subspace spanned only by $\{\varphi_\mu, \varphi_\lambda\}$ is sufficient because
    \begin{equation}
        \begin{split}
            x_1 &= \varphi_\mu \\
            x_2 &= \varphi_\lambda +b \varphi_\mu ^2
        \end{split}
    \end{equation}
    Therefore, we should at least expect our network to find one three-dimensional Koopman invariant subspace with a linear decoder or a two-dimensional Koopman invariant subspace with a nonlinear decoder. In Fig~\ref{fig:nonlinear_ode}, we show the results of trained PIKNs with and without a linear decoder in two columns respectively. The networks have accurately identified the Koopman eigenvalues and eigenfunctions\footnote{We don't expect eigenfunctions to be exactly $[x_1, x_1^2, x_2-bx_1^2]$ because each one is still a valid eigenfunction (associated with the same eigenvalue) when being multiplied by some factors. Therefore, this statement is correct up to some constant scale.} within the corresponding invariant subspace. Both models exhibit accurate predictive performance. 
    Note that these two networks are both trained in a purely physics-informed manner, indicating there is no need for simulation data. Specific details about the network architecture, training procedure and other extensive studies are provided in the Appendix~\ref{sec:simple_ode_appendix}.

\subsection{Heat equation}
\label{sec:heat_eqn}

    \begin{figure}[t!]
        \centering
        \includegraphics[width=0.85\textwidth]{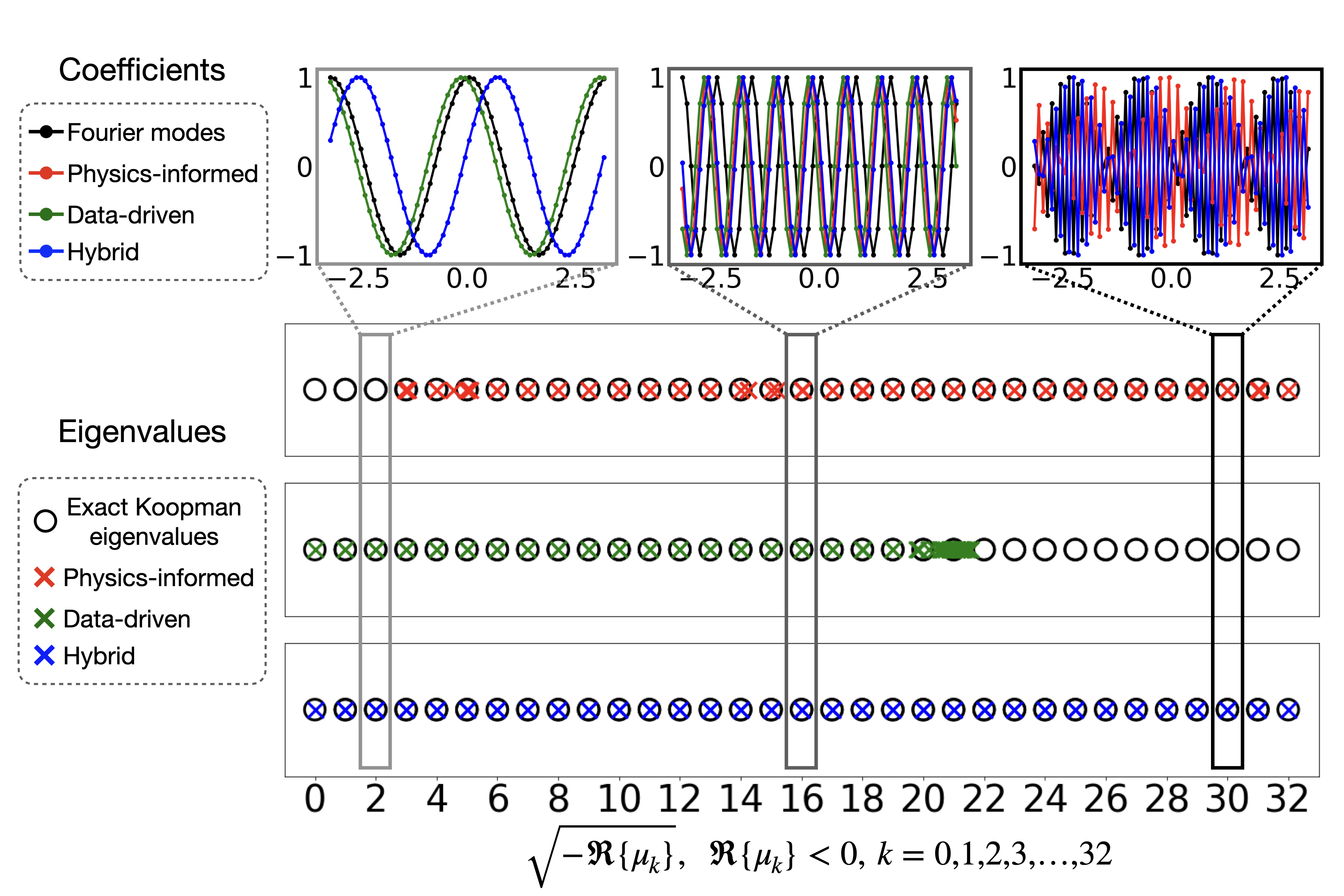}
        \vspace{-.1in}
        \caption{The eigenvalues (with negative real part) of the matrix $L$ from different neural networks are plotted along with the exact, discrete-time eigenvalues of the heat equation at the bottom. The top row shows the coefficients of the linear transformation corresponding to the selected eigenvalues.}
        \label{fig:heat_eigs_dft}
    \end{figure}
    
    \begin{figure}[t!]
        \centering
        \includegraphics[width=\textwidth]{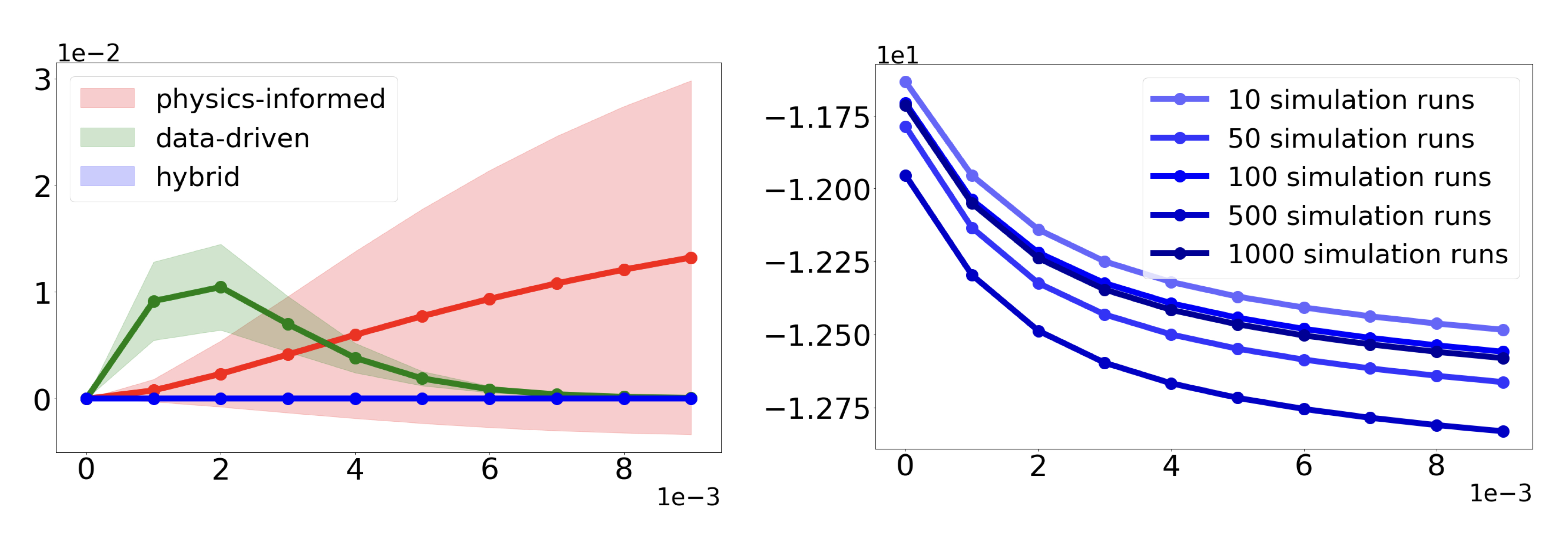}
        \vspace{-.3in}
        \caption{The left plot shows the predictive power of three different neural networks. The prediction task is evaluated on $1000$ newly simulated test trials. The shaded region highlights one standard deviation from the mean squared error. The right plot shows the error curves of $5$ hybrid models. These models differ on the number of simulated trajectories used for training. And the error is visualized in a logarithmic scale of base $10$. Both plots are $10$-step forward predictions with $\Delta t=0.001$.}
        \label{fig:heat_err}
    \end{figure}
    
    The first PDE we consider is the one-dimensional heat equation:
    \begin{equation}
        u_t = u_{xx}, \ \ \ \ x\in(-\pi, \pi)
    \end{equation}
    with periodic boundary conditions. Using Fourier transform, it can be shown that
    discrete-time eigenvalues are \cite{evans2010partial}
    \begin{equation}
        \mu_k = -k^2, \ \ \ k=0, \pm 1, \pm 2, \dots
    \end{equation}
    To approximately represent the trajectories $u$ and the collocations $u_{t}$, we discretize the spatial domain with $n=64$ equally spaced grids. As a consequence, the value of $k$ only ranges from $-32$ to $31$. Therefore, we expect our network to at least mimic a discrete Fourier transform and its inverse transform, identifying the right eigenvalues after training. In addition, we study the effects of data integration. More specifically, we compare the results obtained from the networks that is purely physics-informed, purely data-driven and a mixture of both which we call it hybrid model. We use sums of harmonic functions with random coefficients as collocations for which the analytic spacial derivatives are available. More details of the experimental setup can be found in the Appendix~\ref{sec:heat_equation_appendix}. 
    
    Indeed, Fig~\ref{fig:heat_eigs_dft} shows that the transformation coefficients collide with the frequencies of the corresponding Fourier modes. The phase difference is expected because Discrete Fourier transformation is not unique for diagonalization of the heat equation. The networks nearly identify all the correct eigenvalues of the heat equation, however, the purely physics-informed network fails to discover the low-frequency modes. In addition, we observe it identifies an eigenvalue with a small positive real part (around $0.00005$) which is not visualized in this plot. On the contrary, the purely data-driven network misses the high-frequency modes but all identified eigenvalues have negative real parts; the hybrid model presents the most satisfying accuracy among all. 
    
    The reason behind this result might be the multiscale feature of the system. Namely (i) for data-driven model, we use simulation data which are integrated over time. Slow dynamics (which are associated with the slow frequencies) are more persistent which dominate in the overall loss. (ii) For physics-informed model, however, we use the spatial derivatives of the states which are more sensitive to the high-frequency (transient) modes. (iii) The hybrid model leverages these two time scales and achieves a nice balance for identifying all eigenvalues. This important result suggests using collocations to inform the model of the aspects of the phenomenon that is missing in the data.
    
    Fig~\ref{fig:heat_err} shows the error curves of the networks trained in different ways. The left plot shows physics-informed network is good at short-term predictions while the data-driven network is more promising in long-term predictions. The hybrid network merges the best part of these two and consistently offer the most accurate predictions. We also find from the right plot that the performance of the hybrid model is not sensitive to the number of simulation data used for training. More importantly, training with the largest number of simulation data does not necessarily gives the best prediction results. This is probably because in principle the physics-informed training is sufficient for finding all modes, adding simulation data can only help better discover the persistent modes. This suggests that if the network is physics-informed, data demand from simulation is low although data integration is beneficial for identifying a more accurate model.

\subsection{Burger's equation}

    \begin{figure}[t!]
        \centering
        \includegraphics[width=\textwidth]{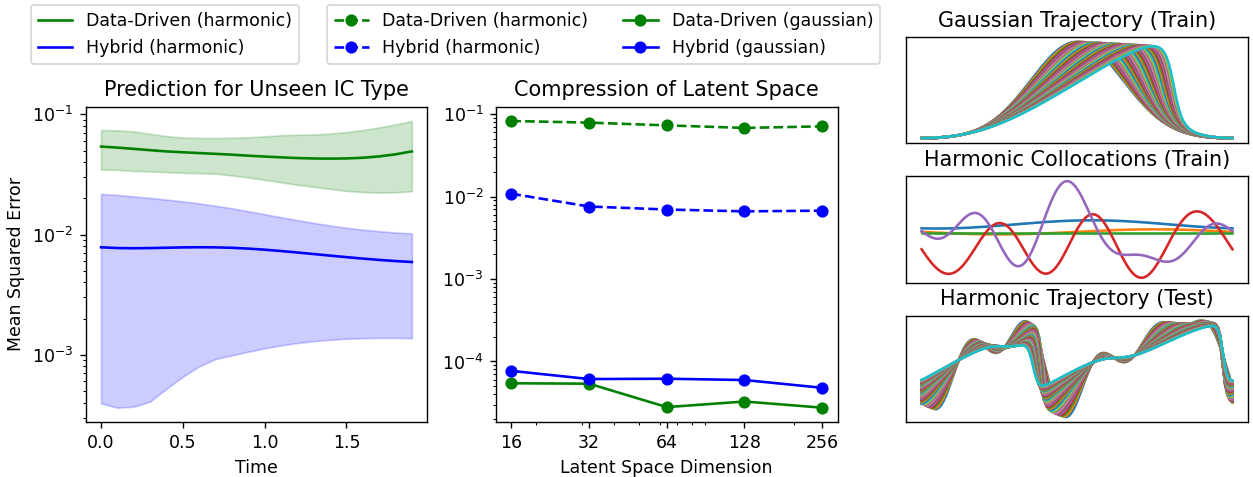}
        \caption{The right plots give examples of data snapshots used: trajectories with bell-curve ICs (top) and harmonic ICs (bottom). The middle-right pane shows harmonic collocations used by hybrid models in addition to the snapshots. The left plot compares the prediction errors (MSE) of two models, data-driven and hybrid with 128-dimensional latent spaces, on harmonic initial conditions that were not present in the trained data. The shaded regions represent 95\% confidence intervals based on 100 test trajectories. The middle plot shows the prediction error (MSE) on each type of test data for a variety of models with different latent-space sizes. We see that the use of harmonic collocations significantly improves the model performance on unseen harmonic ICs without increasing the errors on bell-curve ICs.}
        \label{fig:burgers_compression}
    \end{figure}
    For the next example, we consider the nonlinear PDE known as the Burger's equation
    \begin{equation}
        \label{eq:burgers}
        u_t = -uu_x + \nu u_{xx}, \ x \in (-\pi, \pi)
    \end{equation} 
with periodic boundary conditions. We choose a small value of $\nu=0.01$ such that the solution is advection-dominated, for which the linearization is more challenging \cite{peherstorfer2022breaking,kutz2016dynamic}.

We show that one can use collocations to improve model's performance on types of initial conditions that are missing in the available training data. For that we train two models. The data-driven model only uses 1024 trajectories with bell-curve initial conditions (ICs) (we provide an example at the top-right frame of the Figure~\ref{fig:burgers_compression}). The hybrid model additionally observes 80000 harmonic collocations formed by summing first 10 sinusoidal modes with random coefficients (Figure~\ref{fig:burgers_compression}, middle-right frame), for which we evaluate $u_t$ analytically using Equation~\ref{eq:burgers}. Next, we evaluate the performance of both models using unseen trajectories with both harmonic and bell-curve ICs, 100 trajectories each. We observe that the Hybrid model predicts the sinusoidal trajectories ~10 times better than the data-driven one (Figure~\ref{fig:burgers_compression}, left frame, shown for the 128-dimensional latent-space model). Since neither models had any trajectories of that type in its training set we conclude that the difference in performance comes from using harmonic collocations. We also note that better performance of the hybrid model on harmonic ICs does not come at an expense of worse performance on bell-curve ICs, as shown in the central frame of Figure~\ref{fig:burgers_compression}. This evidence suggests that one can improve a model's extrapolation power  by supplementing its training with sufficiently diverse set of collocations, especially when additional simulations are expensive to obtain but the collocations are cheap to generate. The details of the network's architecture and training procedure are provided in the Appendix \ref{sec:burgers_new_appendix}.

Finally, we highlight a remarkable compressibility of the latent space of PIKN models: a model with a latent space size 16 predicts only two times worse, by MSE, than a model with 256-dimensional latent space (central frame of Figure~\ref{fig:burgers_compression}). It provides an empirical evidence that using a large latent space is not necessary for achieving good practical performance, albeit one may not discover Koopman eigenfunctions. To illustrate it, we compare our results with that of the exact Koopman eigenvalues and eigenfunctions for the Burger's equation, that can be found using Cole-Hopf transformation in the appendix \ref{sec:coleHope}. The learned transformations are not Cole-Hopf; indeed, Cole-Hopf transformations may not be the only one to linearize Burger's equation, similar to the problem arised in \cite{gin2021deep}. However, the linear dynamics in the latent space still give good prediction, especially after adding physics-informed regularization, as evidenced by Figure~\ref{fig:burgers_compression}. Ultimately, high compressibility enables using PIKN for compressed sensing and online control applications. \cite{korda2018linear} 


\section{Discussion and Conclusion}
\label{sec:conclusion}
In this work, we presented an effective deep learning framework for identifying Koopman eigenvalue and eigenfunction pairs for reconstructing high-dimensional nonlinear dynamics. In order to validate our method, we carefully went through three examples on which the analytical form of the Koopman eigen-decomposition can be derived.
To the best of our knowledge, this is the first work that leverages knowledge of physics to improve the performance of auto-encoder-based Koopman learning.  
Our results show that (i) by imposing the Lie equation via soft penalty one can reduce the need of large training data-sets as being required in previous works; (ii) since the framework is under the scope of operator learning, our model can be used for future state predictions on unseen initial states, and (iii) using appropriate collocations one can improve the prediction accuracy on those unseen states by assimilating the knowledge of the dynamics into the model. \\
This work also suggests a number of future research directions. For example, many nonlinear systems may have different Koopman eigen-decompositions over different domains\cite{page2019koopman}. Being able to identify the boundaries of these domains would greatly expand the scope of applications of this approach. It is also interesting to see how these model can benefit the control problems, as in the real world, it is the ultimate goal of studying nonlinear dynamics.

\newpage
\bibliographystyle{plain}
\bibliography{ref}


\newpage
\appendix
\section{Appendix}

\subsection{Experimental setups}
\label{sec:exp_setups}
For all networks in the experiments, $L$ is set to a zero matrix initially and all other parameters are randomly initialized using the default initializer of Pytorch. The exponential linear unit (ELU) is used as the nonlinear activation function, unless stated otherwise, as we expect the transformations in PIKN to be relatively smooth. 

\subsubsection{Simple nonlinear system with discrete spectrum}
\label{sec:simple_ode_appendix}
In this example, we have trained two different types of PIKN: a PIKN with a linear decoder and a PIKN with a nonlinear decoder. In both experiments, $1000$ collocation points were uniformly sampled from $[-1, 1]\times[-1, 1]$ for training, an Adam optimizer with a learning rate of $1e-4$ has been applied. The total number of training epochs is set to $50000$ and the weights in the loss function are set to $\omega_1 = \omega_2 = 1$. The encoder part of both architectures are the same: a 2-layer fully-connected neural network with hidden layer containing $50$ neurons. The major differences between the two architectures are as follows:
\begin{itemize}
    \item The linear decoder is simply a linear layer without a bias term whereas the nonlinear decoder is symmetric to the encoder: a 2-layer fully-connected neural network with hidden width $50$.
    \item For the PIKN with a linear decoder, we have a three-dimensional latent space whereas for the PIKN with a nonlinear decoder, it is two-dimensional.
\end{itemize}

\begin{table}[t!]
    \centering
    \begin{tabular}{|c|c|c|}
        \hline
         & \textbf{PIKN(linear decoder)} & \textbf{PIKN(nonlinear decoder)} \\
        \hline
        \textbf{Experiment 1} & \makecell{$\mu_1=-0.10000689$, $\mu_2=-0.19914131$, \\ $\mu_3=-0.9996788$} & \makecell{$\mu_1=-0.09756267$, \\ $\mu_2=-0.99973810$} \\
        \hline
        \textbf{Experiment 2} & \makecell{$\mu_1=-0.10009335$, $\mu_2=-0.19947967$, \\ $\mu_3=-1.0004123$} & \makecell{$\mu_1=-0.09644943$, \\ $\mu_2=-0.99867420$} \\
        \hline
        \textbf{Experiment 3} & \makecell{$\mu_1=-0.10005733$, $\mu_2=-0.19892442$, \\ $\mu_3=-1.0003881$} & \makecell{$\mu_1=-0.09592330$, \\ $\mu_2=-0.99915651$} \\
        \hline
        \textbf{Experiment 4} & \makecell{$\mu_1=-0.10008135$, $\mu_2=-0.20019206$, \\ $\mu_3=-0.9986593$} & \makecell{$\mu_1=-0.09784412$, \\ $\mu_2=-1.00298023$} \\
        \hline
        \textbf{Experiment 5} & \makecell{$\mu_1=-0.09990316$, $\mu_2=-0.20018657$, \\ $\mu_3=-0.9991975$} & \makecell{$\mu_1=-0.09837312$, \\ $\mu_2=-1.00187280$} \\
        \hline
        \textbf{Experiment 6} & \makecell{$\mu_1=-0.09997816$, $\mu_2=-0.19966313$, \\ $\mu_3=-1.0005931$} & \makecell{$\mu_1=-0.09764695$, \\ $\mu_2=-1.00517617$} \\
        \hline
        \textbf{Experiment 7} & \makecell{$\mu_1=-0.10009032$, $\mu_2=-0.19991782$, \\ $\mu_3=-1.0000535$} & \makecell{$\mu_1=-0.09866422$, \\ $\mu_2=-0.99622254$} \\
        \hline
        \textbf{Experiment 8} & \makecell{$\mu_1=-0.09998395$, $\mu_2=-0.19948480$, \\ $\mu_3=-0.9996783$} & \makecell{$\mu_1=-0.09856838$, \\ $\mu_2=-1.00204348$} \\
        \hline
        \textbf{Experiment 9} & \makecell{$\mu_1=-0.09996726$, $\mu_2=-0.19905518$, \\ $\mu_3=-1.0005406$} & \makecell{$\mu_1=-0.09929386$, \\ $\mu_2=-0.99679565$} \\
        \hline
        \textbf{Experiment 10} & \makecell{$\mu_1=-0.10004932$, $\mu_2=-0.19994377$, \\ $\mu_3=-0.9993069$} & \makecell{$\mu_1=-0.09756234$, \\ $\mu_2=-1.00279380$} \\
        \hline
        \textbf{summary} & \makecell{$\mathbf{\mu_1=-0.10\pm6.22e-04}$, \\ $\mathbf{\mu_2=-0.20\pm 4.37e-04}$, \\ $\mathbf{\mu_3=-1.00\pm 5.99e-05}$} & \makecell{$\mathbf{\mu_1=-0.10\pm 2.74e-03}$, \\ $\mathbf{\mu_2=-1.00\pm 9.68e-04}$} \\
        \hline
    \end{tabular}
    \vspace{+.1in}
    \caption{Eigenvalues identified by PIKNs at all training runs. The first column represents the PIKN with a linear decoder and the second column is the one with a nonlinear decoder.}
    \label{tab:robustness}
\end{table}

\subsubsection{Heat equation}
\label{sec:heat_equation_appendix}

In the PIKN architecture for Heat equation, encoder and decoder are both linear and the latent dimension is set to $64$, the same as the number of spatial grids. Number of training epochs is $100000$ and an Adam optimizer has been applied for the training algorithm. In this set of experiments, we use an adaptive learning rate: initially set to $0.01$, it keeps decreasing by a factor of $0.5$ if no improvements are made over the recent $5000$ epochs, until it hits the minimal value of $0.000001$. \\
With the network architecture and training parameters fixed, we train it in three different ways. Namely, we set different values for the weights $\omega_1, \omega_2, \omega_3, \omega_4$ in the loss function, leading to three different training regimes:
\begin{itemize}
    \item $\omega_1=0.0001, \omega_2=1, \omega_3=0, \omega_4=0$: the network is purely physics-informed.
    \item $\omega_1=0, \omega_2=0, \omega_3=1, \omega_4=1$: this corresponds to a purely data-driven learning.
    \item $\omega_1=0.0001, \omega_2=1, \omega_3=1, \omega_4=1$: this represents the scenario where we train a physics-informed network with data integration. For simplicity, we call it hybrid training.
\end{itemize}
Notice that the value of $\omega_1$ is set on a different order compared to other weights, it is because this loss term involves calculation of numerical derivatives which usually has a greater amplitude. This is a well-known issue for physics-informed neural network and has been thoroughly studied. Adaptive re-weighting schemes were proposed to fix it \cite{wang2021understanding, maddu2021inverse, zubov2021neuralpde}. In our case, however, we find that choosing a fixed, small value $\omega_1=0.0001$ is sufficient for achieving a fast convergence\footnote{Another way to get around this is to learn the pseudo-inverse of $L$ instead. Then the loss term reads $\|\phi(\mathbf{x_i})-L^\dagger \nabla\phi(\mathbf{x_i})\cdot \mathbf{f}(\mathbf{x_i})\|$. In that case $\omega_1$ and $\omega_2$ can be both set to $1$ because the two loss terms are approximately on the same scale.}. \\
For the physics-informed learning, we create $1000$ trial functions $\{u^{(1)}, u^{(2)}, \dots, u^{(1000)}\}$ for training purpose. Each trial function $u^{(j)}$ is a superposition of the Fourier modes that satisfy the periodic boundary conditions. In our case, this amounts to using $\sin{(kx)}$ and $\cos{(kx)}$ for $k=0, 1, 2, \dots$ as basis functions. The value of $k$ is restricted to be no greater than $32$ due to the choice of our grid spatial resolution. The spatial derivatives of the trail functions $\{u_{xx}^{(1)}, u_{xx}^{(2)}, \dots, u_{xx}^{(1000)}\}$ are calculated using numerical spectral method. \\
In the data-driven regime, we use simulation data obtained from a solver based on spectral method. We run the simulation with $1000$ different initial states. For each run, the initial state of $u$ is obtained through the same procedure as we obtain the trial functions $u^{(k)}$. Then we sample snapshots of the state $u$ with a temporal gap $\Delta t=0.01$ for $5$ steps (i.e. $p=5$). \\
For the hybrid training regime, both of the above two data-sets are used. However, to study the effects of the amount of simulation data, we conduct $10$, $50$, $100$, $500$, $1000$ simulation runs in $5$ separate groups of experiments.

\subsubsection{Burger's Equation}
\label{sec:burgers_new_appendix}
In the PIKN architecture for Burger's equation, the encoder is implemented as a a feed-forward network with the size of input layer set to be equal 128 (the number of spatial grid-points). It has two hidden layers with 512 neurons. The size of the output layer of the encoder -- the size of the latent dimension -- varies from 16 to 256 to study the latent space compressibility (see Figure~\ref{fig:burgers_compression}). The decoder's architecture mirrors the architecture of the encoder with the sizes of the inputs and outputs switched. An Adam optimizer has been applied for $500$ epochs for training. We use an adaptive learning rate: initially set to $10^{-4}$, it keeps decreasing by a factor of $0.5$ if no improvement has been made over the recent $20$ epochs, until it hits the minimal value of $10^{-7}$. \\
Each network was trained in two different ways. Namely, we set different values for the weights $\omega_1, \omega_2, \omega_3, \omega_4$ in the loss function, leading to two different training regimes:
\begin{itemize}
    \item $\omega_1=0, \omega_2=0, \omega_3=1, \omega_4=1$. These settings lead to a purely data-driven learning; it corresponds to green lines on Figure~\ref{fig:burgers_compression}.
    \item $\omega_1=1, \omega_2=1, \omega_3=1, \omega_4=1$. These settings represent a model that uses both data trajectories and collocations (a hybrid model); it corresponds to blue lines on Figure~\ref{fig:burgers_compression}.
\end{itemize}
We sample $80000$ functions $\{u^{(1)}, u^{(2)}, \dots, u^{(80000)}\}$ to use them as collocations. Each function $u^{(j)}$ is a superposition of the Fourier modes that satisfy the periodic boundary conditions. We call them "harmonic" initial conditions, five examples of which are displayed on the middle-right pane of Figure~\ref{fig:burgers_compression}. We used $\sin{(kx)}$ and $\cos{(kx)}$ for $k=0, 1, 2, \dots$ evaluated on $x \in [-\pi; \pi]$-interval as basis functions. The value of $k$ is restricted to be no greater than $10$. To evaluate $u_t$ at each collocation point we evaluated the spatial derivatives $\{(u_{x}^{(i)}, u_{xx}^{(i)})\}_{i=1}^{80000}$ using a numerical spectral method. 

For training trajectories, we use simulation data obtained from a solver base on spectral method. We run the simulation with $1024$ different initial states. Each initial state of $u$ is a Gaussian (bell-) curve with mean 0 and a randomly generated variance $\sigma \sim U(0.1,1)$, evaluated on $[-\pi; \pi]$-interval. Then we sample snapshots of the state $u$ with a temporal gap $\Delta t=0.1$ for $20$ steps (i.e. $p=20$). An example trajectory is displayed on the top-right pane of Figure~\ref{fig:burgers_compression}.

To evaluate and compare the performance of both models we use 200 trajectories: 100 trajectories with harmonic ICs and 100 trajectories with Gaussian ICs generated exactly in the same way as described above. Then we sample snapshots of the state $u$ with a temporal gap $\Delta t=0.1$ for $20$ steps (i.e. $p=20$).
The results are aggregated in Table~\ref{tab:burgers_performance} and visualized on Figure~\ref{tab:burgers_performance}.

\begin{table}[t!]
    \centering
    \include{performance_table}
    \caption{Performance measures for Burger's experiment depending on the dimensions of the latent space (Figure~\ref{fig:burgers_compression})}
    \label{tab:burgers_performance}
\end{table}

\subsubsection{Hardware}
\label{sec:hardware}
    All experiments were computed on a \texttt{slurm}-allocation that had 2 CPUs of an \texttt{Intel(R) Xeon(R) CPU E5-2630 v4 @ 2.20GHz}, 16 GB of memory, and one \texttt{Tesla K80} GPU. The experiments were implemented using \texttt{Python 3.9.12} and \texttt{PyTorch 1.11.0} that was using the GPU for compute. 

\subsection{Additional discussions}
\label{sec:add_results}

\subsubsection{Potential pitfalls of using nonlinear transformations}
\label{sec:add_limits}
For the example of ODE with discrete spectrum, we have run the training algorithm for $10$ times for both architectures and the results are robust in the sense that the identified eigenvalues are all centered around the real ones we derived analytically, which are presented in Table~\ref{tab:robustness}. One can see our PIKNs faithfully recover the desired Koopman eigenvalues. The one with the linear decoding provides slightly more robust results, indicating the benefits of adding more known constraints. \\
We notice that, however, the PIKN with nonlinear decoder doesn't always identify the eigenvalue $\mu_1 = -0.1$. If we significantly change the initialization of the network parameters, sometimes other values emerge. This indicates other transformations exist to linearize the dynamics and reconstruct the state variables. As an example, one can easily check $\varphi_{\mu}^\beta = x_1^\beta$ for all $\beta \in \mathbb{N}$ are all valid Koopman eigenfunctions associated with eigenvalues $\mu\beta$ and can be used for reconstruction. The lesson here is that by using a nonlinear decoder, we not only increase the flexibility of the Koopman operator theory framework, but also dramatically increase the searching space, which may lead to different learning outcomes.

\subsubsection{Unknown parameters in physical systems}
\label{sec:unknown_params}
\begin{table}[]
    \centering
    \begin{tabular}{cccc}
    \toprule
    (\#collocations,\\ \#snapshots) & $\mu=-0.1$ &  $\lambda=-1$ & eigenvalues=(-1, -0.1) \\
    \midrule
    $(0, 1000)$ & -- & -- & $(-0.962 \pm 0.118, -0.096 \pm 0.282)$ \\
    $(250, 750)$ & $-0.099 \pm 0.002$ & $-0.980 \pm 0.033$ & $(-0.974 \pm 0.043, -0.099 \pm 0.001)$ \\
    $(500, 500)$ & $-0.099 \pm 0.000$ & $-0.991 \pm 0.005$ & $(-0.985 \pm 0.010, -0.099 \pm 0.001)$ \\
    $(750, 250)$ & $-0.099 \pm 0.003$ & $-0.972 \pm 0.043$ & $(-0.951 \pm 0.073, -0.096 \pm 0.010)$ \\
    \bottomrule
    \end{tabular}
    \caption{Networks are trained with different combinations of data-sets shown in the first column. The first row represents data-driven training and others are hybrid PIKN. The identified system parameters $\mu$ and $\lambda$ and eigenvalues of Koopman operator are shown, respectively, in second, third, and fourth column.}
    \label{tab:unknown_physics}
\end{table}

In this experiment, we demonstrate another advantage of PIKN in comparison to data-driven only approaches. We demonstrate that PIKN can leverage partial knowledge of physics, e.g. when some parameters of the system are unknown and even estimate those missing parameters. 
To illustrate this, we consider the dynamics of the form of Eq. \ref{eq:simpleODE} but we let $\mu$ and $\lambda$ to be unknown parameters, which can be treated as trainable parameters with random initialization. We use different combinations of training data-sets and each model is trained for $10$ times with different random initializations for the unknown parameters.
The difference of data-sets is related to the combination of snapshots (obtained by simulator) versus collocation points (sampled from appropriate function space with no requirement of simulation).
The results are shown in Table~\ref{tab:unknown_physics}: all hybrid models successfully identified Koopman eigenvalues with higher accuracy than the data-driven model (first row). Note that, the data driven model does not involve a parameter estimation procedure and thus does not provide any knowledge of the physical parameters. On the other hand, the hybrid models are able to solve for the unknown parameters $\mu$ and $\lambda$ correctly by filling gaps in the knowledge of physics with data, effectively operating as a simple model-discovery tool. The experiment shows that incorporating a physics-informed loss is beneficial even when only partial knowledge of physics is accessible to the practitioner.  

\subsubsection{Alternative approach to enforce linearity}
In fact, minimizing $\|L \phi(\mathbf{x}) - \nabla \phi(\mathbf{x}) \cdot \mathbf{f}(\mathbf{x})\|$ is not the only way to enforce linearity. Take ODE as an example, the decoder reads 
\begin{equation}
    \hat{\mathbf{x}}(t)=\psi(\mathbf{z}(t))
\end{equation}
This implies 
\begin{equation}
    \frac{d\hat{\mathbf{x}}}{dt}=\nabla \psi(\mathbf{z}) \cdot (L\mathbf{z})
\end{equation}
Therefore, minimizing $\|\nabla \psi(\mathbf{z}) \cdot (L\mathbf{z}) - \mathbf{f}(\hat{\mathbf{x}})\|$ is an alternative way to enforce linearity. This alternative approach is also consistent with 'future state prediction' loss in the work of \cite{lusch2018deep}. The difference between these two ways is whether the linear constraint is applied to the encoder or decoder. In practice, however, we find using either form of the linearity loss or use both of them all work well and does not lead to significantly different results.

\subsubsection{Cole-Hopf transfomration and comparison with exact Koopman eigenvalues and eigenfunctions}
\label{sec:coleHope}

\begin{figure}[t!]
        \centering
        \includegraphics[width=0.9\textwidth]{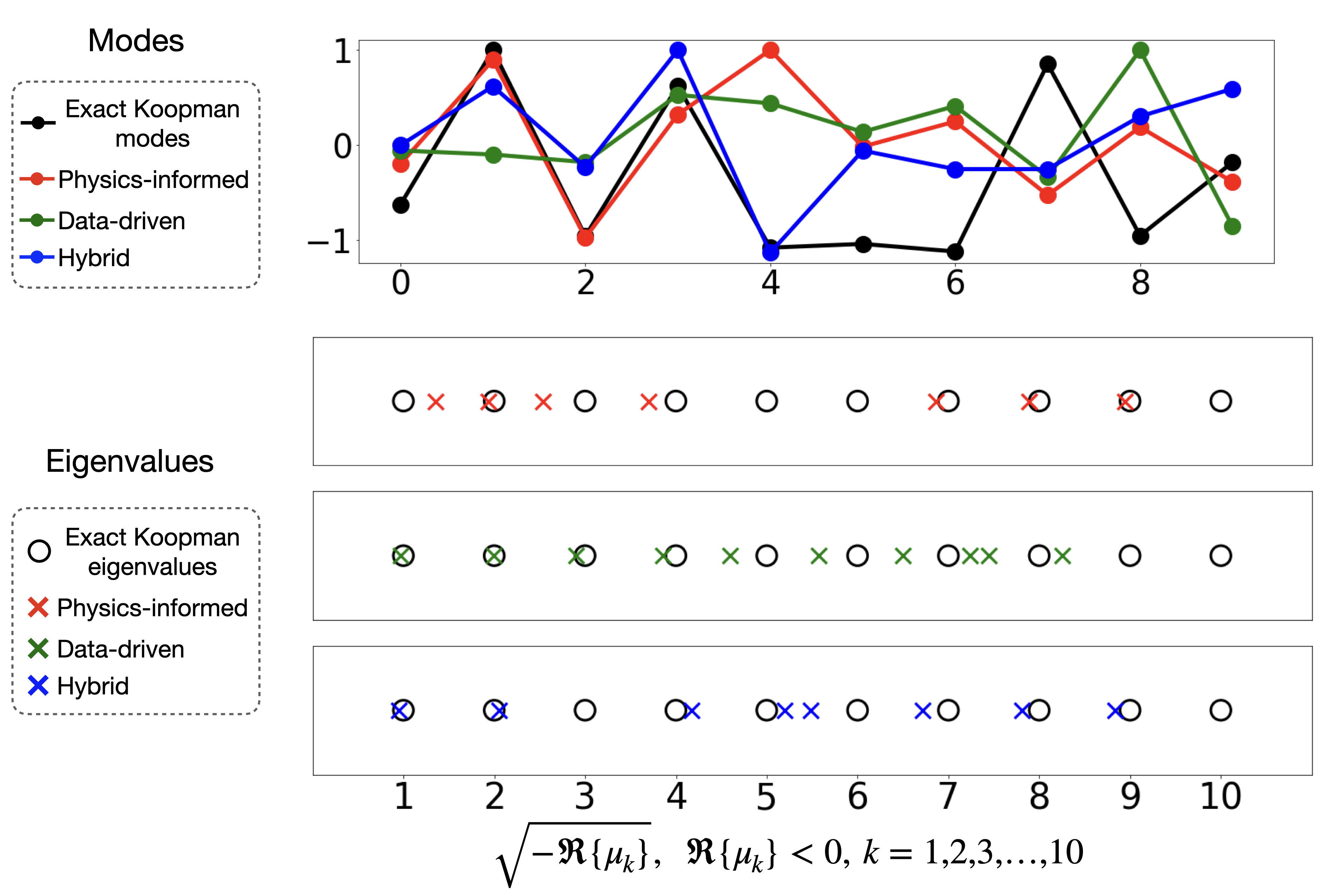}
        \caption{The first row shows the identified modes (or latent representation) of a sampled test input from different network architectures along with the exact Koopman modes. The second row shows eigenvalues (with negative real part) from different neural networks along with the exact, discrete-time eigenvalues of the Burger's equation identified through Cole-Hopf.}
        \label{fig:burgers_eigs}
    \end{figure}

It is known that the Burger's equation can be linearized through Cole-Hopf transformation which was discovered in 1950 and 1951 by Eberhard Hopf\cite{hopf1950partial} and Julian Cole\cite{cole1951quasi}, independently of one another, and later noticed by Kutz et al \cite{nathan2018applied} and exploited by many others\cite{page2018koopman, balabane2021koopman}. The Cole-Hopf transformations are defined as:
    \begin{equation}
        \label{eq:cole-hopf}
        \begin{split}
            & u := C(v) = -\frac{2v_x}{v}\\
            & v := H(u) = \frac{e^{-\frac{1}{2}\int_{-\pi}^{x}u(s, t)ds}}{\int_{-\pi}^{\pi}e^{-\frac{1}{2}\int_{-\pi}^{x}u(s, t)ds}dx}
        \end{split}
    \end{equation}
    If $u(x, t)$ satisfies the Burger's equation~\ref{eq:burgers} with a homogeneous boundary condition $u(-\pi, t)=u(\pi, t)=0$ and an initial condition $u(x, 0) = u_0(x)$, then $v(x, t) = H(u(x, t))$ solves the heat equation:
    \begin{equation}
        \begin{split}
            & v_t = v_{xx}, \ x\in (-\pi, \pi) \\
            & v_x(-\pi, t) = v_x(\pi, t) = 0 \\
            & v(x, 0) := v_0(x) = H(u_0(x))
        \end{split}
    \end{equation}
    The solution of it, by using a standard separation of variables approach, can be derived as 
    \begin{equation}
        v(x, t) = C_0 + \sum_{k=1}^{\infty}C_k \cos(kx) e^{-k^2 t}
    \end{equation}
    where $C_0 = \frac{1}{2\pi}\int_{-\pi}^{\pi}v_0(x)dx$ and $C_k = \frac{1}{\pi}\int_{-\pi}^{\pi}v_0(x)\cos(kx)dx$. \\
    Conversely, it is also shown in \cite{balabane2021koopman} that if $v(x, t)$ solves the above heat equation and further satisfies 
    \begin{equation}
        \begin{split}
            & v(x, t) > 0, \ x\in (-\pi, \pi), t > 0 \\
            & \int_{-\pi}^{\pi}v(x, t)dx = 1
        \end{split}
    \end{equation}
    then $u(x, t) = C(v(x, t))$ is also a solution for the original Burger's equation. To satisfy these constraints, we can require $C_0 = \frac{1}{2\pi}$ and $\sum_{k=1}^{\infty}|C_k| < \frac{1}{2\pi}$. Therefore, we use the following procedure to generate data:
    \begin{itemize}
        \item Choose a number of modes $K$ (i.e. $C_k = 0$ for $k > K$).
        \item Create a list of $\{\tilde{C}_1, \tilde{C}_2, \cdots, \tilde{C}_K\}$ with each $\tilde{C}_k$ uniformly sampled from $(-1, 1)$.
        \item Obtain a number $\alpha$ uniformly sampled from $(0, 1)$.
        \item Create a list of $\{C_1, C_2, \cdots, C_K\}$ with each $C_k = \frac{\alpha\tilde{C}_k}{\sum_{k=1}^K|\tilde{C}_k|}$ (so that $\sum_{k=1}^K|C_k|=\alpha < 1$)
        \item Generate $v(x, t) = \frac{1}{2\pi} + \sum_{k=1}^{K}C_k \cos(kx) e^{-k^2 t}$.
        \item Obtain $u(x, t) = C(v(x, t))$.
    \end{itemize}
    This data generation process guarantees the existence of at least one transformation (Cole-Hopf transformation followed by a Fourier transformation) that linearizes the nonlinear Burger's equation with eigenvalues $\mu_k = -k^2\ (k=1, 2, ..., K^2)$. In our experiments, we set $K=10$. \\
    The architecture used for this experiment consists of a nonlinear encoder and a nonlinear decoder. Both encoder and decoder are 3-layer fully connected neural network with hidden width $512$. The latent dimension is set to the same as the number of modes $K=10$. The training parameters (learning rates and number of training epochs) are set to be the same as in the heat equation experiments. And the number of spatial grids is set to 512.\\
    We also study the three different learning regimes, with different combinations of $\omega_1, \omega_2, \omega_3$ and $\omega_4$. The only difference from the setup of the heat equation experiments is that $\omega_1$ is set to $0.01$. \\
    For physics-informed learning, we create $1000$ trial functions $\{u^{(1)}, u^{(2)}, \dots, u^{(1000)}\}$ for training. Each trial function $u^{(j)}$ is a snapshot at a random time point $t^{(j)} \in [0, 0.1)$ of the $u(x, t)$ generated from the above procedure and projected onto the spatial grids. To generate $\{u_x^{(1)}, u_x^{(2)}, \dots, u_x^{(1000)}\}$ and $\{u_{xx}^{(1)}, u_{xx}^{(2)}, \dots, u_{xx}^{(1000)}\}$, we use $u=C(v)=-\frac{2v_x}{v}$, so that
    \begin{equation}
        \begin{split}
            & u_x = 2(\frac{v_x}{v})^2 - 2\frac{v_{xx}}{v} \\
            & u_{xx} = 6\frac{v_x v_{xx}}{v^2} - 2\frac{v_{xxx}}{v} - 4(\frac{v_x}{v})^3
        \end{split}
    \end{equation}
    where $v_x$, $v_{xx}$ and $v_{xxx}$ are easy to derive. \\
    For data-driven learning, we also obtain data from the above procedure. We sample snapshots of the state $u(x, t)$ with a temporal gap $\Delta t=0.02$ for $5$ steps. For the hybrid regime, both types of data are used. \\
    In fig~\ref{fig:burgers_eigs}, the identified eigenvalues for Burger's equation of each training strategy are plotted along with the exact Koopman eigenvalues. 
    The hybrid model, similar to the results of heat equation, seems to work slightly better for combination of low- and high- frequencies and reveal a more appealing spectrum, compared to data-driven or physics-informed regime alone (although significant improvement is not identified). 
    This indicates the learned transformations are not exactly Cole-Hopf. However, in terms of prediction, we report significant improvement and generalization when unseen initial conditions are used as test data.\\

\end{document}